\documentclass[sigconf,screen]{acmart}
\AtBeginDocument{%
  }

\usepackage{amsmath}

\usepackage{amssymb}
\usepackage{algorithmic}
\usepackage{algorithm}
\usepackage{array}
\usepackage{textcomp}
\usepackage{stfloats}
\usepackage{url}
\usepackage{verbatim}
\usepackage{graphicx}
\usepackage{epsfig}

\usepackage{multirow}
\usepackage{booktabs}
\usepackage{caption}
\usepackage{subcaption}
\usepackage{makecell}

\setcopyright{acmlicensed}
\copyrightyear{2026}
\acmYear{2026}
\acmDOI{XXXXXXX.XXXXXXX}
\acmConference[MM '26]{Proceedings of the 34rd ACM International Conference on Multimedia}{November 10–14,
  2026}{Rio de Janeiro, Brazil}
\acmISBN{978-1-4503-XXXX-X/2018/06}

\renewcommand\footnotetextcopyrightpermission[1]{}

\begin{document}

\title{ResetEdit: Precise Text-guided Editing of Generated Image via Resettable Starting Latent}


\author{Hanyi Wang}
\affiliation{%
  \institution{Shanghai Jiao Tong University}
  \country{China}
  }
\email{why_820@sjtu.edu.cn}

\author{Han Fang}
\affiliation{%
  \institution{University of Science and Technology of China}
  \country{China}
}
\email{fanghan@ustc.edu.cn}


\author{Zheng Wang}
\affiliation{%
  \institution{Wuhan University}
  \country{China}
  }
\email{wangzwhu@whu.edu.cn}

\author{Shilin Wang}
\affiliation{%
  \institution{Shanghai Jiao Tong University}
  \country{China}
  }
\email{wsl@sjtu.edu.cn}

\author{Ee-Chien Chang}
\affiliation{%
  \institution{National University of Singapore}
  \country{Singapore}
}
\email{changec@comp.nus.edu.sg}









\begin{abstract}

    Recent advances in diffusion models have enabled high-quality image generation, leading to increasing demand for post-generation editing that modifies local regions while preserving global structure. Achieving such flexible and precise editing requires a high-quality starting point, a latent representation that provides both the freedom needed for diverse modifications and the precision required for fine-grained, region-specific control. However, existing inversion-based approaches such as DDIM inversion often yield unsatisfactory starting latents, resulting in degraded edit fidelity and structural inconsistency. Ideally, the most suitable editing anchor should be the original latent used during the generation process, as it inherently captures the scene’s structure and semantics. Yet, storing this latent for every generated image is impractical due to massive storage and retrieval costs. To address this challenge, we propose ResetEdit, a proactive diffusion editing framework that embeds recoverable latent information directly into the generation process. By injecting the discrepancy between the clean and diffused latents into the diffusion trajectory and extracting it during inversion, ResetEdit reconstructs a resettable latent that closely approximates the true starting state. Additionally, a lightweight latent optimization module compensates for reconstruction bias caused by VAE asymmetry. Built upon Stable Diffusion, ResetEdit integrates seamlessly with existing tuning-free editing methods and consistently outperforms state-of-the-art baselines in both controllability and visual fidelity.
    
\end{abstract}




\maketitle

\section{Introduction}

Diffusion models have become the dominant approach for high-quality image generation, supporting a wide range of creative applications \cite{ho2020denoising, song2020denoising, rombach2022high}. Beyond one-shot generation, users may require post-generation editing of diffusion-generated images, that is, refining synthetic outputs by modifying specific regions while preserving the overall layout, illumination, and style. For example, a designer may adjust a diffusion-generated rendering by replacing a component without altering the global composition or visual tone. This post-generation scenario highlights a critical requirement: flexible editability, namely fine-grained, component-level control that remains consistent with user intent.

\begin{figure}[t!]
  \centering
  \includegraphics[width=\linewidth]{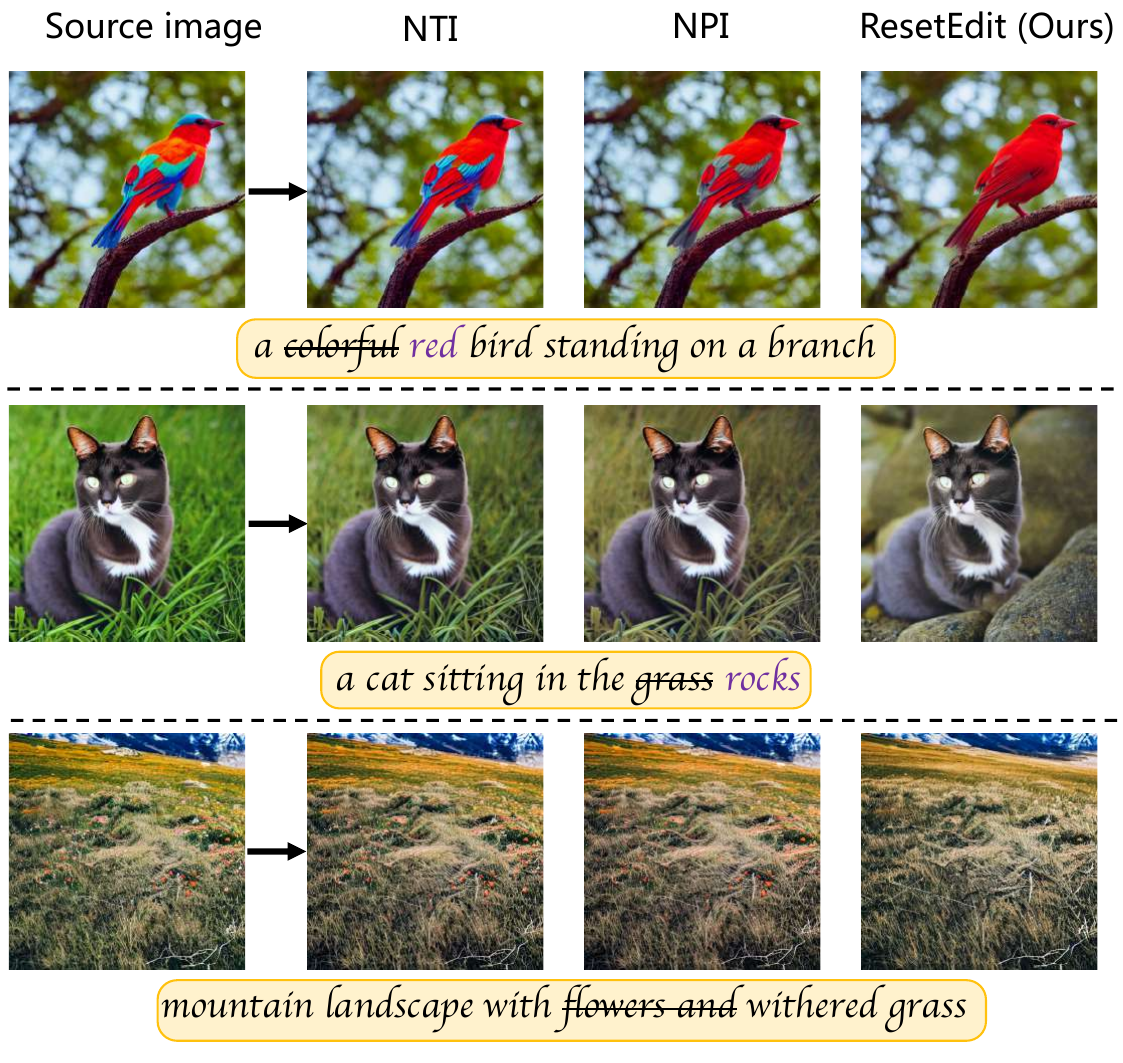}
  \caption{Editing results across NTI \cite{mokady2023null}, NPI \cite{miyake2025negative}, and our ResetEdit. Under the same editing prompts, NTI and NPI fail to faithfully realize the requested changes or introduce unwanted alterations, whereas ResetEdit reconstructs a reliable starting latent that enables accurate and localized edits.}
  \label{fig:fig_1}
\end{figure}

Achieving such high-quality post-generation editing fundamentally relies on the ability to revert a generated image to an editable state, a latent representation from which the image can be faithfully reconstructed and selectively modified. In practice, this is typically achieved through diffusion inversion, where the generated image is mapped back into the latent space as the starting point for re-editing. The most common approach is DDIM inversion \cite{song2020denoising}, which estimates a pseudo initial latent by reversing the deterministic diffusion process. Building upon this idea, recent variants such as Null-text Inversion (NTI) \cite{mokady2023null} optimize the null-conditioning embedding to better reconstruct the input image, while Negative-prompt Inversion (NPI) \cite{miyake2025negative} reformulates NTI into a closed-form, training-free solution. Despite these advances, such inversion-based pipelines often fall short in practice. The recovered latent does not always correspond to a truly editable state, where requested changes may not be faithfully realized, and unintended alterations frequently appear in non-target regions, degrading structural consistency and edit fidelity (see Fig.~\ref{fig:fig_1}, the inversion-based methods, NTI and NPI, struggle to correctly modify color, background, or remove objects, producing unsatisfactory editing).


We argue that, for effective post-generation editing, the ideal starting point should coincide with the original latent used to generate the image. This latent inherently encodes the global structural prior and semantic composition that guided the generation process, making it the most reliable anchor for subsequent edits. Restoring such a latent would allow precise, instruction-aligned modifications without disturbing the overall layout or visual coherence. A straightforward solution might be to store the original starting latent for every generated image. However, this approach quickly becomes impractical at scale, as it imposes significant storage and retrieval overhead, especially for large-scale diffusion deployments where millions of images are produced. This observation motivates a new question: can we recover the original starting latent $x_T$ directly from a given generated image, without any explicit storage, while retaining its reliability as an editable anchor for high-fidelity re-editing?


To answer this question, we propose a proactive solution that transforms latent recovery from a passive inversion process into an active embedding framework. The key idea is to encode recoverable information about the original starting latent directly within the diffusion trajectory during generation. Specifically, we imperceptibly inject the discrepancy between the original starting latent \(x_T\) and its diffused counterpart \(x_0\) into the diffused latent. At inversion time, this embedded signal is extracted to reconstruct a starting latent that closely approximates \(x_T\) with negligible error. In this way, our method provides a reliable anchor that enables high-fidelity and semantically aligned edits.

However, realizing this idea is nontrivial and introduces two major challenges.

\noindent\textbf{(1) Tradeoff between imperceptibility and recoverability.} The embedded signal must remain imperceptible to avoid degrading perceptual quality or disturbing the diffusion trajectory, yet it must be reliably retrievable during inversion. Since the residual is high-dimensional and structurally aligned with the latent itself, naive embedding can distort the generated image or destabilize sampling.

\noindent\textbf{(2) VAE Encoder–decoder asymmetry.} As embedding and extraction occur in latent space, the imperfect correspondence between the VAE encoder and decoder introduces reconstruction bias. This asymmetry perturbs the recovered latent and shifts it away from the true starting state, weakening its reliability as an editing anchor.



To overcome these challenges, we introduce ResetEdit, a fundamentally new diffusion editing framework built around a resettable starting latent. First, to resolve the long-standing tension between imperceptibility and recoverability, we devise a residual injection network that encodes the discrepancy between the original and diffused latents as an invisible yet retrievable signal, akin to a self-embedded “edit key”. The discrepancy is compressed through a vector-quantized variational autoencoder (VQ-VAE), which transforms it into compact, semantically structured codebook indices. These indices are subtly fused into the latent trajectory during forward diffusion, ensuring that the signal remains imperceptible to both human observers and the model’s sampling dynamics. During inversion, a dedicated decoder reconstructs this signal to accurately recover the true starting latent, effectively “resetting” the generation state with near-lossless precision. Second, to tackle the intrinsic encoder–decoder asymmetry of diffusion VAEs, we introduce a latent optimization module that adaptively refines the re-encoded latent through gradient-based alignment. This refinement step compensates for non-invertible distortions accumulated along the encoding–decoding path, significantly improving recovery fidelity and stability. Together, these designs enable ResetEdit to transform latent reconstruction from a post-hoc correction into an integrated, proactive mechanism.


In summary, our contributions are as follows:

\begin{itemize} \item We propose ResetEdit, a novel proactive diffusion editing framework featuring a resettable starting latent. By embedding the latent discrepancy directly into the forward diffusion trajectory and retrieving it during inversion, ResetEdit transforms latent recovery from a passive reconstruction step into an active generative mechanism, enabling faithful reconstruction of the original latent for high-fidelity editing.

\item We design a latent optimization strategy that compensates for VAE encoder–decoder asymmetry, refining the re-encoded latent through adaptive alignment and further enhancing the fidelity and stability of the reconstructed starting point.

\item Extensive experiments demonstrate that ResetEdit produces more faithful starting latents and is fully compatible with existing editing approaches, consistently improving controllability and performance across multiple benchmarks. \end{itemize}

\section{Related Work}

\subsection{Diffusion Models for Image Generation and Editing}

Diffusion models have emerged as a dominant paradigm for high-quality image generation by reversing a noise process through iterative denoising. DDPM \cite{ho2020denoising} establishes the foundation, while DDIM \cite{song2020denoising} enables deterministic sampling and inversion. Classifier-free guidance \cite{ho2022classifier} further improves conditional generation. Building on these advances, large-scale text-to-image models such as DALL·E 2 \cite{ramesh2022hierarchical}, Imagen \cite{saharia2022photorealistic}, and Stable Diffusion \cite{rombach2022high} have achieved remarkable performance. In particular, Stable Diffusion leverages a latent VAE space for efficient and scalable generation.

Beyond generation, diffusion models also support text-guided image editing \cite{zhan2022multimodal}. We focus on tuning-free approaches \cite{meng2021sdedit, hertz2022prompt, tumanyan2023plug, cao2023masactrl, liu2024towards}, which avoid retraining. Representative methods include Prompt-to-Prompt (P2P) \cite{hertz2022prompt}, Plug-and-Play (PnP) \cite{tumanyan2023plug}, and MasaCtrl \cite{cao2023masactrl}, which respectively rely on cross-attention control, feature modulation, and mask-guided attention to enable editing.

\subsection{Diffusion Inversion Methods}

Diffusion inversion aims to recover an editable latent representation from a given image, enabling faithful reconstruction and structure-preserving editing. Most approaches rely on DDIM inversion \cite{song2020denoising}, which works well in unconditional settings but suffers from large reconstruction errors under classifier-free guidance \cite{ho2022classifier}.

To address this issue, various methods have been proposed. Null-text inversion (NTI) \cite{mokady2023null} optimizes null-text embeddings but incurs high computational cost. Negative-prompt inversion (NPI) \cite{miyake2025negative} and its extension ProxNPI \cite{han2023improving} provide efficient inference-time alternatives with improved accuracy. EDICT \cite{wallace2023edict} achieves exact inversion via dual diffusion paths at the cost of doubled computation. DDPM inversion \cite{huberman2024edit} reconstructs images using forward noise but degrades editing quality, while direct inversion \cite{ju2023direct} improves reconstruction by substituting latents but limits flexibility.

In contrast, we posit that high-fidelity editing should rely on the true starting latent of the generation process rather than approximations from inversion. Based on this insight, we propose a proactive alternative that enables precise recovery of the starting latent with negligible error.
\begin{figure*}[t!]
  \centering
  \includegraphics[width=0.9\textwidth]{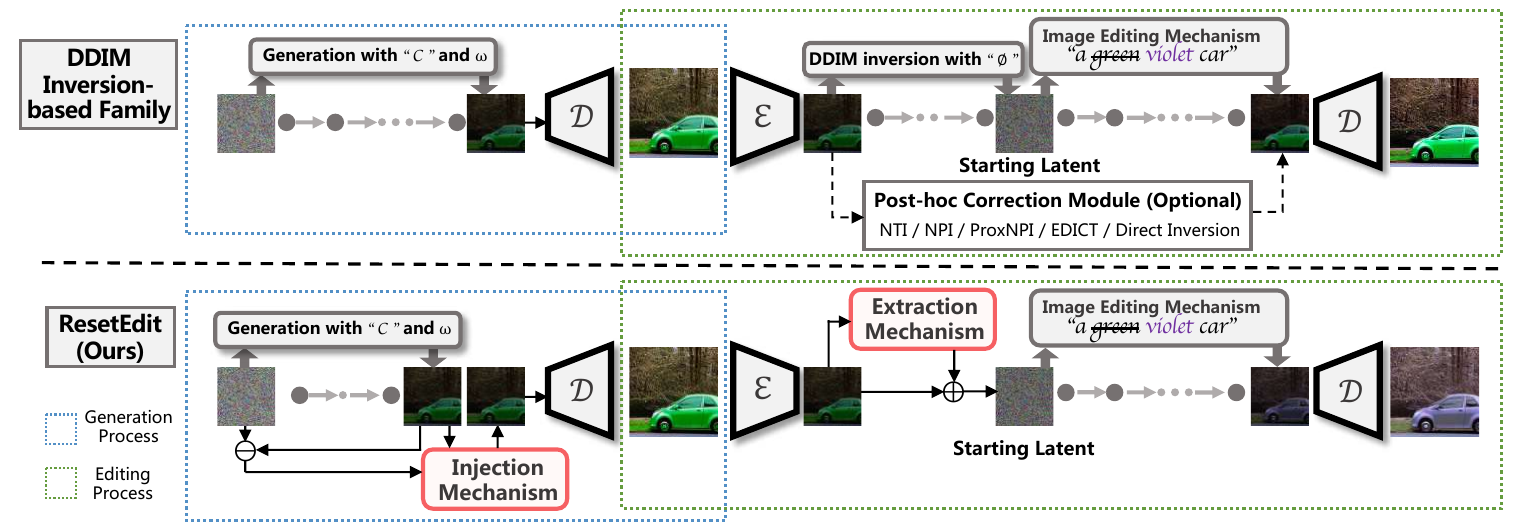}
 \caption{Comparison between DDIM inversion-based editing (top) and ResetEdit (bottom). In the DDIM-based family, NTI, NPI, and ProxNPI primarily mitigate \textit{Condition Drift}, while EDICT, DDPM inversion, and Direct Inversion primarily aim to reduce \textit{Estimation Error}. However, all remain constrained by the asymmetry between forward and backward processes. In contrast, ResetEdit proactively embeds the discrepancy during generation and later extracts it to reconstruct a faithful starting latent, providing a reliable anchor for precise and semantically aligned editing.}
 \label{fig:fig_2}
\end{figure*}

\section{The Proposed ResetEdit}

\subsection{Analysis and Solutions}
As discussed above, precise and flexible image editing relies on a suitable starting latent. Ideally, this latent should be the exact one used to generate the image to be edited. However, recovering such a latent through inversion is non-trivial. Since existing methods predominantly rely on standard DDIM inversion, we begin this section by analyzing the limitations of DDIM inversion, and then introduce our approach for obtaining a more reliable starting latent.

\subsubsection{Limitations of DDIM inversion.}

The DDIM forward process at timestep $t$ is defined as:
\begin{equation}
x_{t-1} = \gamma_t x_t + \varphi_t \{\epsilon_\theta(x_t, t, \emptyset) + w \left[ \epsilon_\theta(x_t, t, C) -\epsilon_\theta(x_t, t, \emptyset)\right] \}, 
\label{eq:eq_1}
\end{equation}

\noindent
where \( \gamma_t = \sqrt{\alpha_{t-1}/\alpha_t} \) and \( \varphi_t = -\sqrt{\alpha_{t-1}(1-\alpha_t)/\alpha_t} + \sqrt{1 - \alpha_{t-1}} \). Here, \( \{\alpha_t\}_{t=0}^T \) denotes a monotonically increasing noise schedule.

The corresponding inversion step is approximated as:
\begin{align}
x'_t \approx \frac{1}{\gamma_t}x_{t-1} - \frac{\varphi_t}{\gamma_t}\epsilon_\theta(x_{t-1}, t, \emptyset).
\label{eq:eq_2}
\end{align}
based on the assumption that \(\epsilon_\theta(x_t, t, \emptyset) \approx \epsilon_\theta(x_{t-1}, t, \emptyset)\) \cite{wallace2023edict}.

In practical editing scenarios, only the final image is available, and the original prompt $C$ and guidance scale $w$ are typically unknown. As a result, inversion can only proceed under unconditional guidance (i.e., with ``$\emptyset$''), leading to a mismatch between the forward and backward processes. This discrepancy introduces inversion error, which can be decomposed as: 
\begin{equation}
    \begin{split}
x'_t - x_t & = \frac{\varphi_t}{\gamma_t} \Big\{\underbrace{w\big[ \epsilon_\theta(x_t, t, C) - \epsilon_\theta(x_t, t, \emptyset)\big]}_{\text{Condition Drift}} \\
& + \underbrace{\epsilon_\theta(x_t, t, \emptyset) - \epsilon_\theta(x_{t-1}, t, \emptyset)}_{\text{Estimation Error}} \Big\}.
    \end{split}
\label{eq:eq_8}
\end{equation}
These errors stem from two main sources: (1) \textit{Condition Drift}, the mismatch in prompt and classifier-free guidance between sampling and inversion; and (2) \textit{Estimation Error}, approximation error $\epsilon_\theta(x_t, t, \emptyset)$ with $\epsilon_\theta(x_{t-1}, t, \emptyset)$. Such errors accumulate over the inversion trajectory, ultimately resulting in an inaccurate recovery of the original latent. 


\subsubsection{Limitations of inversion-based approaches.}
As analyzed above, DDIM inversion inevitably suffers from \textit{Condition Drift} and \textit{Estimation Error} (Eq.~\ref{eq:eq_8}), which accumulate along the trajectory and distort the recovered latent. A number of follow-up works have attempted to address these issues. On the one hand, methods such as NTI~\cite{mokady2023null}, NPI~\cite{miyake2025negative}, and ProxNPI~\cite{han2023improving} mainly focus on alleviating condition drift, either by optimizing null-text embeddings or by reformulating prompt conditioning. On the other hand, approaches including EDICT~\cite{wallace2023edict}, DDPM inversion~\cite{huberman2024edit}, and Direct Inversion~\cite{ju2023direct} primarily seek to reduce estimation errors by modifying the inversion trajectory or exploiting auxiliary noise information. 

Despite these advances, all such methods share a common limitation: they are fundamentally post-hoc corrections within the DDIM inversion framework. Since the forward and backward processes are inherently asymmetric—due to unknown prompts, classifier-free guidance, and approximation assumptions—these approaches can only partially mitigate errors. Consequently, none of them is able to faithfully recover the original starting latent, which remains the most reliable anchor for high-fidelity re-editing.

\begin{figure*}[t!]
  \centering
  \includegraphics[width=\textwidth]{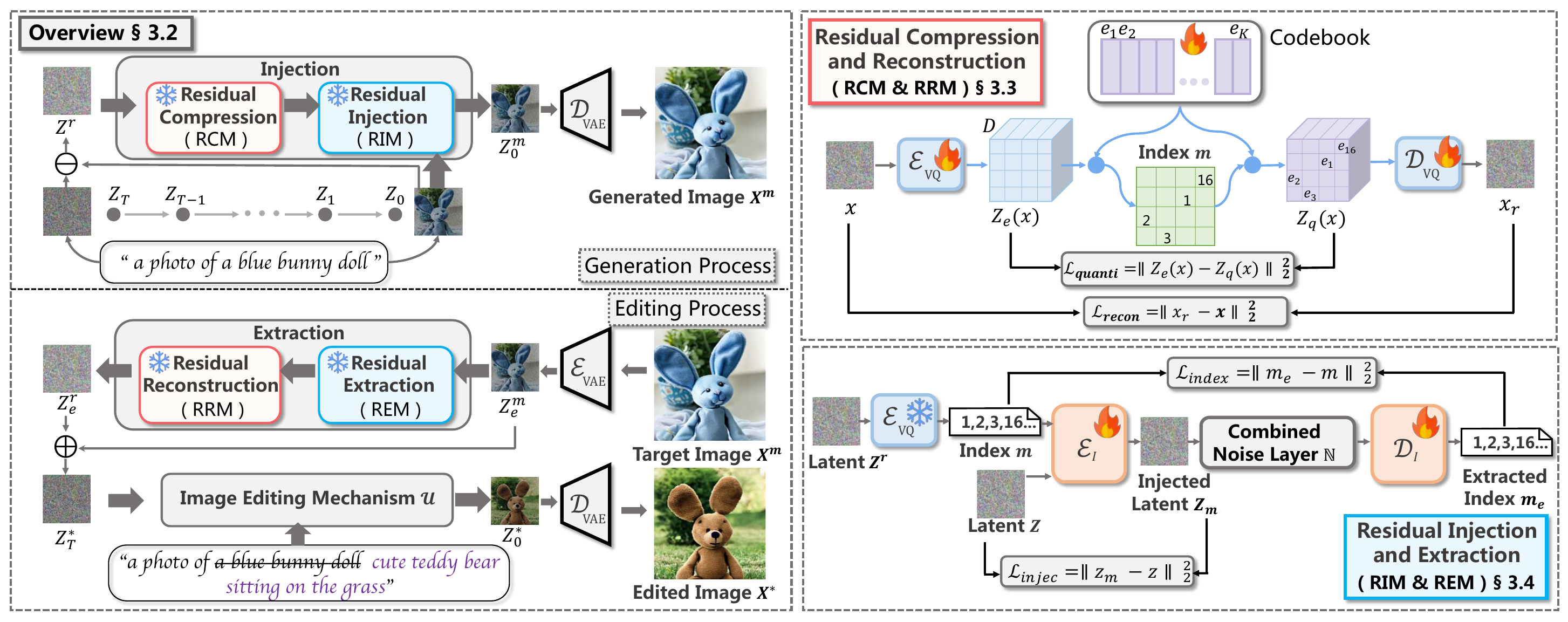}
  \caption{The framework of our proposed ResetEdit, which comprises two key modules: Residual Compression and Reconstruction Modules (RCM and RRM), and Residual Injection and Extraction Modules (RIM and REM). During image generation, RCM compresses the latent discrepancy, which is then injected by RIM. In the editing phase, REM first extracts the embedded residual, and RRM reconstructs an accurate starting latent for subsequent precise image editing.}
  \label{fig:fig_3}
\end{figure*}

\subsubsection{Proactive resettable starting latent}

Given the inherent limitations of inversion-based methods, post-hoc corrections cannot fully eliminate inversion errors. We therefore propose a proactive strategy: instead of reconstructing the starting latent after generation, we explicitly embed the predictable discrepancy into the diffused latent during forward sampling. This signal is carefully designed to be imperceptible, preserving the visual fidelity of the generated image while remaining reliably retrievable. During inversion, the embedded residual is extracted and added back to the estimated latent, enabling near-lossless recovery of the original starting point. In this way, our approach establishes a resettable starting latent that provides a faithful and robust basis for subsequent editing. For an intuitive comparison, Figure~\ref{fig:fig_2} contrasts conventional DDIM-based editing with our proposed method.

\subsection{Framework Overview}
By integrating the aforementioned components, we construct the complete \textbf{ResetEdit} framework, as illustrated in Figure~\ref{fig:fig_3}. The framework consists of four key components: a diffusion model \( \mathcal{G} \) coupled with a pre-trained VAE \( \{\mathcal{E}_{\text{VAE}}, \mathcal{D}_{\text{VAE}}\} \), Residual Compression and Reconstruction Modules (RCM and RRM), Residual Injection and Extraction Modules (RIM and REM), and an image editing mechanism $\mathcal{U}$, which modulates attention maps to enable controllable editing. The RCM and RRM are responsible for compressing the residual into a compact index and reconstructing it during inversion. The RIM and REM handle the injection and extraction of the residual signal. RCM/RRM and RIM/REM are trained independently and remain fixed during inference.

The workflow of the proposed framework can be described as: For generation process, given an initial Gaussian noise \( z_T \), the diffusion model iteratively refines it over \( T \) denoising steps, conditioned on the text prompt \( C \) and guided by a classifier-free guidance scale \( w \), producing the image latent representation \( z_0 = \mathcal{G}(z_T, T, C, w) \). We then compute the residual latent \( z_r = z_T - z_0 \), which captures the discrepancy between the initial noise and the generated latent. This residual is first passed through the RCM, which compresses \( z_r \) into a discrete representation indexed by a learned codebook, denoted as \( e_r \). Further, the index of the compressed residual, \( m_r \), is used to represent \( e_r \) and is injected into the image latent \( z_0 \) via the RIM, yielding a \textit{residual-carrying image latent} \( z_0^m \). This latent carries the residual information in a visually imperceptible manner, without altering the semantic integrity of the original image. Next, the residual-carrying latent \( z_0^m \) is decoded by the pre-trained VAE decoder to generate the residual-carrying image \( X^m = \mathcal{D}_{\text{VAE}}(z_0^m) \). 

To enable pixel-domain editing, \( X^m \) is re-encoded using the VAE encoder to obtain the latent representation \( z_e^m = \mathcal{E}_{\text{VAE}}(X^m) \), which corresponds to the residual-carrying latent. It is worth noting that the VAE encoder does not serve as an exact inverse of the decoder, which introduces an inherent discrepancy between the re-encoded latent \( z_e^m \) and the original latent \( z_0^m \). To mitigate this reconstruction error, we perform an iterative optimization process on \( z_e^m \) over \( n \) steps to obtain a refined latent representation, denoted as \( z_{\tilde{e}}^{m} \). This optimized latent \( z_{\tilde{e}}^{m} \) is then passed through the REM to recover the injected binary code \( m_r \), and subsequently through the RRM to reconstruct the residual latent \( z_e^r \). By adding this reconstructed residual to the optimized latent, we obtain an accurate approximation of the original noise used in generation: \( z_T^* = z_e^{\prime m} + z_e^r \). Finally, the reconstructed latent \( z_T^* \), together with a user-provided editing prompt \( C^* \), is fed into the image editing mechanism $\mathcal{U}$. $\mathcal{U}$ conducts a guided forward diffusion process conditioned on \( C^* \), resulting in the generation of the edited image \( X^* \).

\subsection{Training of RCM and RRM}

To inject the residual latent representation into the image latent in a visually imperceptible and nearly lossless manner, we first compress the residual information. To this end, RCM and RRM are jointly designed based on the vector-quantized variational autoencoder (VQ-VAE) framework \cite{van2017neural}.

Specifically, RCM compresses the input latent \( x \in \mathbb{R}^{C \times H \times W} \) into a continuous representation \( z_e(x) \in \mathbb{R}^{C' \times  H' \times W'} \) using a encoder \( \mathcal{E}_{\text{VQ}} \). Each vector \( z_e(x)_{i,j} \) is then quantized by mapping it to the nearest codeword in a learned codebook \( e \in \mathbb{R}^{K \times D} \), where \( K \) is the number of codewords and \( D \) is the embedding dimension:
\[
k_{i,j} = \arg\min_{l} \| z_e(x)_{i,j} - e_l \|_2, \quad z_q(x)_{i,j} = e_{k_{i,j}}.
\]
This quantization produces a discrete code \( m_r \in \{1, ..., K\}^{H' \times W'} \), where each element \( k_{i,j} \) is an integer in the range \( 1 \leq k_{i,j} \leq K \), corresponding to the index \( l \) of the nearest codeword in the learned codebook \( e \in \mathbb{R}^{K \times D} \), where \( K \) is the number of codewords and \( D \) is the embedding dimension. This code \( m_r \) is then injected into the image latent. During reconstruction, RRM retrieves the indices \( k_{i,j} \) from \( m_r \), maps them back to their corresponding codewords, and decodes the resulting latent representation via a decoder \( \mathcal{D}_{\text{VQ}} \) to reconstruct the residual latent \( x_r \).

To train the encoder \( \mathcal{E}_{\text{VQ}} \), decoder \( \mathcal{D}_{\text{VQ}} \), and the codebook \( e \), we minimize a loss function that combines both reconstruction and quantization objectives. The overall training objective is defined as:
\[
\mathcal{L} = \mathcal{L}_{\text{recon}} + \beta \mathcal{L}_{\text{quanti}},
\]
where
\[
\mathcal{L}_{\text{recon}} = \| x_r - x \|_2^2, \quad \mathcal{L}_{\text{quanti}} = \| z_e(x) - \text{sg}[z_q(x)] \|_2^2.
\]

Here, \( \mathcal{L}_{\text{recon}} \) encourages accurate reconstruction of the residual latent, and \( \mathcal{L}_{\text{quanti}} \) promotes codebook commitment by aligning encoder outputs with their quantized counterparts. The stop-gradient operator \( \text{sg}[\cdot] \) prevents gradients from propagating into the codebook. The hyperparameter \( \beta \) balances reconstruction fidelity and quantization consistency.

\subsection{Training of RIM and REM}

RIM and REM are designed to inject and recover the discrete residual code \( m_r \) within the image latent in a robust and imperceptible manner. Specifically, the injection network \( \mathcal{E}_w \) takes the image latent \( z \) and an index information \( m \) as input, and outputs an injected latent \( z_m = \mathcal{E}_w(z, m) \). To enhance robustness, a combined noise layer \( \mathcal{N} \) is applied to \( z_m \), yielding a perturbed latent \( \widetilde{z}_m = \mathcal{N}(z_m) \). This noisy latent is then passed through the extraction network \( \mathcal{D}_w \) to recover the index: \( m_e = \mathcal{D}_w(\widetilde{z}_m) \).

We train \( \mathcal{E}_w \) and \( \mathcal{D}_w \) using a composite loss that balances message accuracy and injection imperceptibility:
\[
\mathcal{L} = \mathcal{L}_{\text{index}} + \lambda \mathcal{L}_{\text{injec}},
\]
where
\[
\mathcal{L}_{\text{index}} = \| m_e - m \|_2^2, \quad \mathcal{L}_{\text{injec}} = \| z_m - z \|_2^2.
\]

The index loss \( \mathcal{L}_{\text{index}} \) promotes accurate recovery of the injected code under noise, while the injection loss \( \mathcal{L}_{\text{injec}} \) encourages the injected latent to remain close to the original, preserving visual quality. The weighting factor \( \lambda \) controls the trade-off between robustness and imperceptibility.

\subsection{VAE Optimization}

To mitigate the errors introduced by VAE asymmetry, we perform iterative optimization on the re-encoded latent representation. Using the latent \( z_e^m \) from the VAE encoder as initialization, we optimize it to better approximate the original latent \( z_0^m \). The objective is to minimize the difference between the regenerated image and the original residual-carrying image \( X^m \). At each step, the latent is decoded by the pre-trained VAE decoder \( \mathcal{D}_{\text{VAE}} \), and the reconstruction loss is computed in pixel space as:
\begin{equation}
\mathcal{L}_{\text{opt}} = \left\| \mathcal{D}_{\text{VAE}}(z_{\tilde{e}}^{m}) - X^m \right\|_2^2,
\label{eq:vae_opt_loss}
\end{equation}

\noindent
where \( z_{\tilde{e}}^{m} \) denotes the optimized latent. This process yields a refined representation that improves reconstruction fidelity and supports more accurate residual recovery.

\begin{table*}[ht]
\centering
\resizebox{0.95\textwidth}{!}{
\begin{tabular}{l|ccc|ccc|ccc|ccc}
\toprule

\textbf{Dataset}
& \multicolumn{6}{c|}{\textbf{PIE-Bench}} 
& \multicolumn{6}{c}{\textbf{ImageNet-R-TI2I}} \\

\midrule 
\textbf{Metric} & \multicolumn{3}{c|}{\textbf{CLIP Score} $\uparrow$} & \multicolumn{3}{c|}{\textbf{CLIP Similarity} $\uparrow$} & \multicolumn{3}{c|}{\textbf{CLIP Score} $\uparrow$} & \multicolumn{3}{c}{\textbf{CLIP Similarity} $\uparrow$} \\

\midrule

\textbf{Method} & P2P & PnP & MasaCtrl & P2P & PnP & MasaCtrl & PtP & PnP & MasaCtrl & P2P & PnP & MasaCtrl \\

\midrule

DDIM Inv. \cite{song2020denoising} 
& 0.2860 & 0.2942 & 0.2830 & 0.0774 & 0.1414 & 0.1189 & 0.2711 & 0.2825 & 0.2663 & 0.0960 & 0.1914 & 0.0767 \\

NTI \cite{mokady2023null} 
& 0.2966 & 0.3071 & 0.2878 & 0.1617 & 0.1957 & 0.1493 & 0.2897 & 0.3130 & 0.2793 & 0.1900 & 0.2670 & 0.1426 \\

NPI \cite{miyake2025negative} 
& 0.2993 & 0.3065 & 0.2899 & 0.1811 & 0.2127 & 0.1539 & 0.2928 & 0.3086 & 0.2861 & 0.2160 & 0.2686 & 0.1617 \\

ProxNPI \cite{han2024proxedit} 
& 0.2953 & 0.3003 & 0.2867 & 0.1635 & 0.1893 & 0.1375 & 0.2844 & 0.2936 & 0.2698 & 0.1886 & 0.2207 & 0.1191\\

EDICT \cite{wallace2023edict} 
& 0.2811 & 0.2299 & 0.2865 & 0.0954 & 0.1103 & 0.1162 & 0.2654 & 0.2120 & 0.2705 & 0.1231 & 0.1471 & 0.1348 \\

DDPM Inv. \cite{huberman2024edit} 
& 0.2745 & 0.2828 & 0.2747 & 0.0484 & 0.0958 & 0.0457 & 0.2557 & 0.2701 & 0.2543 & 0.0859 & 0.1384 & 0.0797 \\

Direct Inv. \cite{ju2023direct} 
& 0.2895 & 0.2950 & 0.2891 & 0.1165 & 0.0311 & 0.0476 & 0.2753 & 0.2848 &  0.2601 & 0.1353 & 0.1879 & 0.0788 \\

\midrule

ResetEdit (Ours) & \textbf{0.3061} & \textbf{0.3162} & \textbf{0.3070} & \textbf{0.2162} & \textbf{0.2253} & \textbf{0.1575} & \textbf{0.2948} & \textbf{0.3157} & \textbf{0.2868} & \textbf{0.2482} & \textbf{0.2806} & \textbf{0.1640} \\

\bottomrule
\end{tabular}}
\caption{Quantitative comparison of different inversion methods combined with three representative tuning-free editing pipelines (P2P, PnP, and MasaCtrl) on \textit{PIE-Bench} and \textit{ImageNet-R-TI2I}. Results are reported in terms of CLIP Score and CLIP Similarity ($\uparrow$ indicates higher is better). ResetEdit consistently achieves the best performance across all pipelines and benchmarks, demonstrating its effectiveness in reconstructing a reliable starting latent for high-fidelity editing.}
\label{tab:tab_1}
\vspace{-10pt}
\end{table*}
\begin{table*}[h]
\centering
\renewcommand{\arraystretch}{1.1} 
\setlength{\tabcolsep}{3pt}       

\begin{subtable}{\textwidth}
\centering
\resizebox{0.95\textwidth}{!}{
\begin{tabular}{l|ccccccc|cc}
\toprule
\multicolumn{1}{l|}{\textbf{Type}} 
& \multicolumn{7}{c|}{\textbf{Object}} 
& \multicolumn{2}{c}{\textbf{Style}} \\
\midrule
\multicolumn{1}{l|}{\textbf{Sub-type}} 
& \textbf{change object} 
& \textbf{add object} 
& \textbf{delete object} 
& \makecell{\textbf{change attr.}\\\textbf{content}} 
& \makecell{\textbf{change attr.}\\\textbf{pose}} 
& \makecell{\textbf{change attr.}\\\textbf{color}} 
& \makecell{\textbf{change attr.}\\\textbf{material}} 
& \textbf{change background} 
& \textbf{change style} \\
\midrule

DDIM Inv. \cite{song2020denoising} & 0.2783/0.1634 & 0.2889/0.0003 & 0.2821/0.0092 & 0.2967/0.0620 & 0.3148/0.0027 & 0.2886/0.1181 & 0.3032/0.0614 & 0.2766/0.0665 & 0.2842/0.0851 \\

NTI \cite{mokady2023null} & 0.2971/0.2664 & 0.2983/0.0730 & 0.2855/0.0676 & 0.3006/0.1030 & 0.3164/\underline{0.0233} & 0.3012/0.2713 & 0.3027/0.1093 & 0.2941/0.1788 & \underline{0.2935}/\underline{0.1180} \\

NPI \cite{miyake2025negative} & \underline{0.3025}/\underline{0.2817} & \underline{0.2988}/\underline{0.0773} & \underline{0.2864}/\underline{0.0872} & \underline{0.3046}/\underline{0.1267} & 0.3158/0.0227 & \underline{0.3083}/\textbf{0.3298} & \underline{0.3046}/\textbf{0.1151} & \underline{0.2985}/\underline{0.2013} & 0.2907/0.1146 \\

ProxNPI \cite{han2024proxedit} & 0.2972/0.2710 & 0.2963/0.0638 & 0.2857/0.0826 & 0.3003/0.1085 & 0.3154/0.0186 & 0.3048/0.3005 & 0.3008/0.0998 & 0.2927/0.1821 & 0.2843/0.0931 \\

EDICT \cite{wallace2023edict} & 0.2721/0.1733 & 0.2919/0.0403 & 0.2817/0.0409 & 0.2906/0.0667 & 0.3142/0.0145 & 0.2850/0.1564 & 0.2878/0.0498 & 0.2729/0.0927 & 0.2775/0.0770 \\

DDPM Inv. \cite{huberman2024edit} & 0.2637/0.1253 & 0.2883/0.0180 & 0.2798/0.0180 & 0.2874/0.0390 & 0.3149/0.0026 & 0.2755/0.0658 & 0.2849/0.0254 & 0.2652/0.0290 & 0.2716/0.0525 \\

Direct Inv. \cite{ju2023direct} & 0.2850/0.2189 & 0.2952/0.0462 & 0.2816/0.0292 & 0.2982/0.0726 & \underline{0.3186}/0.0221 & 0.2926/0.1854 & 0.2971/0.0766 & 0.2801/0.1043 & 0.2866/0.0910 \\

\midrule
ResetEdit (Ours) & \textbf{0.3235}/\textbf{0.3512} & \textbf{0.3009}/\textbf{0.0932} & \textbf{0.3017}/\textbf{0.1627} & \textbf{0.3141}/\textbf{0.1694} & \textbf{0.3189}/\textbf{0.0454} & \textbf{0.3140}/\underline{0.3018} & \textbf{0.3069}/\underline{0.1142} & \textbf{0.3136}/\textbf{0.2554} & \textbf{0.3037}/\textbf{0.1459} \\

\bottomrule
\end{tabular}}
\end{subtable}

\vspace{0.7em}

\begin{subtable}{\textwidth}
\centering
\resizebox{0.95\textwidth}{!}{
\begin{tabular}{l|ccccccc|cc}
\toprule
\multicolumn{1}{l|}{\textbf{Type}} 
& \multicolumn{7}{c|}{\textbf{Object}} 
& \multicolumn{2}{c}{\textbf{Style}} \\
\midrule
\multicolumn{1}{l|}{\textbf{Sub-type}} 
& \textbf{change object} 
& \textbf{add object} 
& \textbf{delete object} 
& \makecell{\textbf{change attr.}\\\textbf{content}} 
& \makecell{\textbf{change attr.}\\\textbf{pose}} 
& \makecell{\textbf{change attr.}\\\textbf{color}} 
& \makecell{\textbf{change attr.}\\\textbf{material}} 
& \textbf{change background} 
& \textbf{change style} \\
\midrule
DDIM Inv. \cite{song2020denoising} & 0.2865/0.2271 & 0.2996/0.0878 & 0.2849/0.0953 & 0.3023/0.0952 & 0.3084/0.0311 & 0.2967/0.2043 & 0.3031/0.1061 & 0.2929/0.1404 & 0.2933/0.0985 \\

NTI \cite{mokady2023null} & 0.3028/0.2679 & \underline{0.3012}/0.0912 & 0.2940/0.1204 & 0.3041/0.1318 & 0.3154/\underline{0.0371} & \underline{0.3159}/0.3113 & \underline{0.3182}/0.1466 & \underline{0.3084}/0.2372 & \textbf{0.3199}/\underline{0.1812} \\

NPI \cite{miyake2025negative} & \underline{0.3038}/\textbf{0.2904} & 0.3003/\textbf{0.1024} & \underline{0.2961}/\underline{0.1525} & \underline{0.3076}/\textbf{0.1573} & 0.3170/0.0318 & 0.3140/\textbf{0.3504} & 0.3147/\underline{0.1565} & \underline{0.3084}/\underline{0.2523} & \underline{0.3176}/\textbf{0.1821} \\

ProxNPI \cite{han2024proxedit} & 0.2982/\underline{0.2723} & 0.2963/0.0807 & 0.2953/0.1453 & 0.3062/0.1409 & 0.3157/0.0244 & 0.3078/\underline{0.3310} & 0.3078/0.1342 & 0.2982/0.2169 & 0.3008/0.1385 \\

EDICT \cite{wallace2023edict} & 0.2166/0.1525 & 0.2348/0.0771 & 0.2320/0.1043 & 0.2174/0.0764 & 0.2501/\textbf{0.0419} & 0.2139/0.1189 & 0.2520/0.1045 & 0.2160/0.1023 & 0.2446/0.1055 \\

DDPM Inv. \cite{huberman2024edit} & 0.2793/0.1931 & 0.2939/0.0548 & 0.2810/0.0470 & 0.2937/0.0809 & 0.3138/0.0045 & 0.2847/0.1272 & 0.2938/0.0844 & 0.2719/0.0767 & 0.2769/0.0684 \\

Direct Inv. \cite{ju2023direct} & 0.2899/0.2226 & 0.2974/0.0560 & 0.2841/0.0613 & 0.3049/0.0917 & \textbf{0.3180}/0.0156 & 0.3005/0.2025 & 0.3057/0.1009 & 0.2937/0.1477 & 0.2885/0.0921 \\

\midrule
ResetEdit (Ours) & \textbf{0.3050}/0.2642 & \textbf{0.3077}/\underline{0.1020} & \textbf{0.2992}/\textbf{0.1530} & \textbf{0.3154}/\underline{0.1503} & \underline{0.3178}/0.0362 & \textbf{0.3165}/0.3285 & \textbf{0.3185}/\textbf{0.1573} & \textbf{0.3087}/\textbf{0.2529} & 0.3160/0.1805 \\

\bottomrule
\end{tabular}}
\end{subtable}

\vspace{0.7em}

\begin{subtable}{\textwidth}
\centering
\resizebox{0.95\textwidth}{!}{
\begin{tabular}{l|ccccccc|cc}
\toprule
\multicolumn{1}{l|}{\textbf{Type}} 
& \multicolumn{7}{c|}{\textbf{Object}} 
& \multicolumn{2}{c}{\textbf{Style}} \\
\midrule
\multicolumn{1}{l|}{\textbf{Sub-type}} 
& \textbf{change object} 
& \textbf{add object} 
& \textbf{delete object} 
& \makecell{\textbf{change attr.}\\\textbf{content}} 
& \makecell{\textbf{change attr.}\\\textbf{pose}} 
& \makecell{\textbf{change attr.}\\\textbf{color}} 
& \makecell{\textbf{change attr.}\\\textbf{material}} 
& \textbf{change background} 
& \textbf{change style} \\
\midrule
DDIM Inv. \cite{song2020denoising} & 0.2877/0.1960 & 0.2926/0.0174 & 0.3009/0.1069 & 0.2940/0.0563 & 0.3164/0.0102 & 0.2769/0.0386 & \underline{0.3002}/0.0493 & 0.2731/0.0402 & 0.2737/0.0481 \\ 

NTI \cite{mokady2023null} & \underline{0.3035}/0.2768 & \textbf{0.3016}/0.0968 & \underline{0.3017}/0.1775 & \underline{0.2968}/0.0803 & \textbf{0.3193}/0.0398 & \underline{0.2789}/0.0925 & 0.2984/0.0725 & \underline{0.2770}/\underline{0.1011} & \underline{0.2807}/0.0832 \\

NPI \cite{miyake2025negative} & 0.3004/\textbf{0.2998} & 0.2994/0.1020 & 0.2946/\textbf{0.1984} & 0.2938/\textbf{0.1097} & 0.3141/\underline{0.0723} & 0.2786/\textbf{0.1214} & 0.2957/\underline{0.0815} & 0.2737/\textbf{0.1108} & 0.2787/\textbf{0.0876} \\

ProxNPI \cite{han2024proxedit} & 0.2989/\underline{0.2940} & 0.2930/0.0934 & 0.2975/\underline{0.1965} & 0.2899/0.0798 & 0.3148/0.0628 & 0.2752/0.1023 & 0.2943/0.0748 & 0.2702/0.0916 & 0.2726/0.0527 \\

EDICT \cite{wallace2023edict} & 0.2886/0.2359 & 0.2957/0.0632 & 0.2990/0.1350 & 0.2942/0.0866 & 0.3149/0.0260 & 0.2786/0.1003 & 0.2923/0.0645 & 0.2733/0.0962 & 0.2777/0.0760 \\

DDPM Inv. \cite{huberman2024edit} & 0.2616/0.1258 & 0.2902/0.0237 & 0.2829/0.0299 & 0.2887/0.0444 & 0.3152/0.0002 & 0.2709/0.0299 & 0.2856/0.0210 & 0.2660/0.0195 & 0.2705/0.0469 \\
Direct Inv. \cite{ju2023direct} & 0.2832/0.1993 & 0.2929/0.0242 & 0.2988/0.1177 & 0.2946/0.0576 & 0.3157/0.0011 & 0.2776/0.0546 & 0.2978/0.0525 & 0.2695/0.0325 & 0.2727/0.0405 \\

\midrule
ResetEdit (Ours) & \textbf{0.3037}/0.2677 & \underline{0.3012}/0.0825 & \textbf{0.3024}/0.1737 & \textbf{0.2975}/\underline{0.0910} & \underline{0.3186}/\textbf{0.0730} & \textbf{0.2834}/\underline{0.1171} & \textbf{0.3009}/\textbf{0.0927} & \textbf{0.2774}/0.0989 & \textbf{0.2855}/\underline{0.0846} \\

\bottomrule
\end{tabular}}
\end{subtable}

\caption{Results on PIE-Bench across three representative editing pipelines: from top to bottom, (a) P2P, (b) PnP, and (c) MasaCtrl. Each entry reports CLIP Score / CLIP Directional Similarity for all sub-categories. Inv. denotes inversion-based methods, and attr. abbreviates attribute. The best results are highlighted in bold, and the second-best is underlined.}
\label{tab:pie_bench_all}
\vspace{-15pt}
\end{table*}

\section{Experiments}

\subsection{Experimental Setup}

\noindent
\noindent
\textbf{Implementation details.} We conduct experiments using Stable Diffusion (SD) \cite{rombach2022high}, specifically the SD-v2.1 model, generating images of size \(512 \times 512 \times 3\) with latent representations of \(4 \times 64 \times 64\). We evaluate on \textit{PIE-Bench} \cite{ju2023direct} (700 images across 10 editing types) and the \textit{ImageNet-R-TI2I} benchmark from Plug-and-Play (PnP) \cite{tumanyan2023plug} (150 prompt pairs). For discrete residual representation learning, the encoded latent \(z_e\) has size \(512 \times 4 \times 4\), and the codebook contains \(K = 16\) entries of dimension \(D = 1024\). The loss weight is set to \(\beta = 1\), and optimization is performed using Adam with a learning rate of \(3 \times 10^{-4}\) and AMSGrad.  The resulting index map is encoded into a 64-bit representation (\(H' \times W' \times \log_2 K = 64\)) and embedded into the latent space. For RIM and REM, we adopt the MBRS framework~\cite{jia2021mbrs}. The model is trained on 1,000 latent samples generated from the SD-Prompt dataset~\cite{StableDiffusionPrompts}, with randomly sampled binary messages. The noise layer includes Gaussian noise (\(\mu = 0, \sigma = 0.05\)) and Gaussian filtering (\(k = 7\)). We set \(\lambda = 0.1\) and optimize using Adam with a learning rate of \(1 \times 10^{-3}\).


\noindent
\textbf{Evaluation Metrics.} 
Following~\cite{liu2024towards}, we use CLIP Score and CLIP Directional Similarity for editing evaluation. 
Injection performance is measured by bitwise extraction accuracy, and reconstruction quality is evaluated by the mean squared error (MSE) between reconstructed and ground-truth latents.

\noindent
\textbf{Baselines.} We evaluate our method with several representative inversion and editing approaches. For the editing dimension, we adopt three widely used tuning-free editing methods: (i) Prompt-to-Prompt (P2P)\cite{hertz2022prompt}, (ii) Plug-and-Play (PnP)\cite{tumanyan2023plug}, and (iii) MasaCtrl~\cite{cao2023masactrl}. For the inversion dimension, we compare with a comprehensive set of inversion techniques, including: DDIM inversion~\cite{song2020denoising}, Null-text inversion (NTI)~\cite{mokady2023null}, Negative-prompt inversion (NPI)~\cite{miyake2025negative}, ProxNPI~\cite{han2023improving}, EDICT~\cite{wallace2023edict}, DDPM inversion~\cite{huberman2024edit}, and Direct inversion~\cite{ju2023direct}. This setup evaluates how inversion strategies interact with different editing pipelines, demonstrating that ResetEdit provides more accurate and robust starting latents.


\begin{figure*}[t!]
  \centering
  \includegraphics[width=0.9\textwidth]{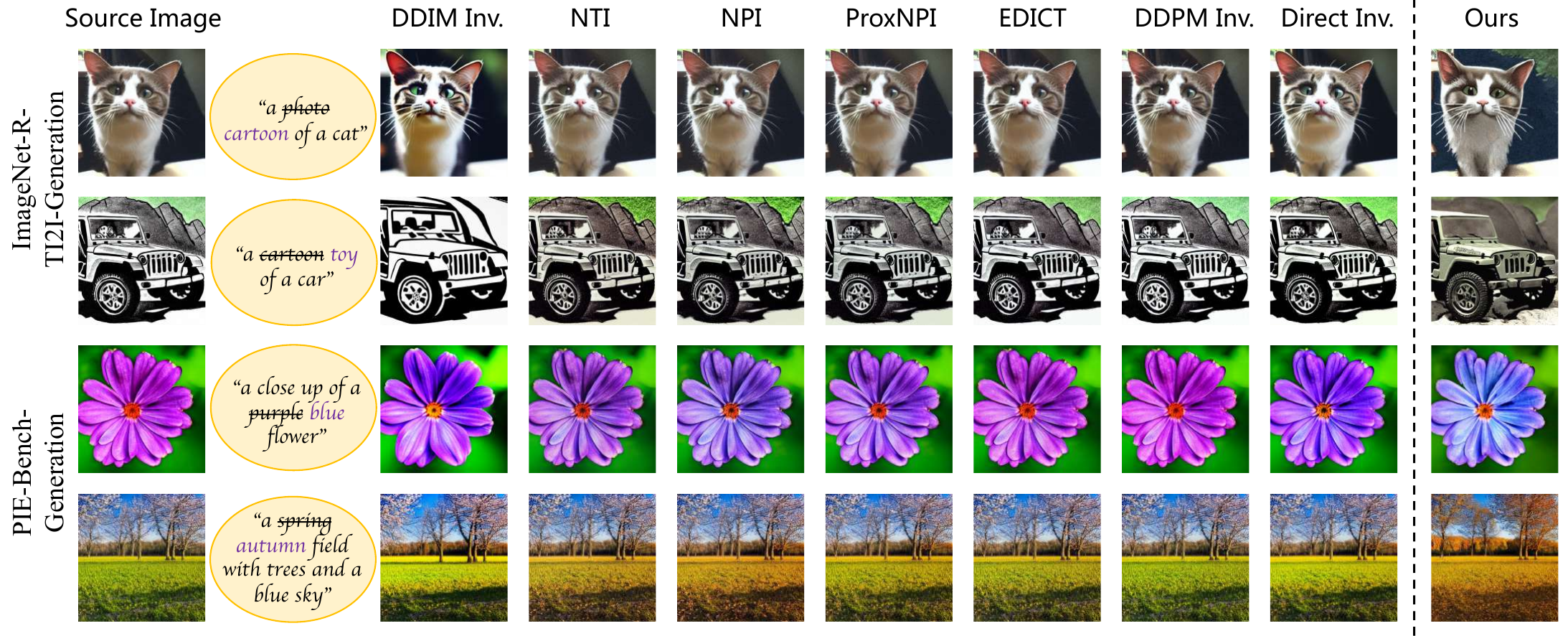}
  \caption{Qualitative comparison of different inversion methods combined with P2P on \textit{ImageNet-R-TI2I} and \textit{PIE-Bench}.}
  \label{fig:fig_4}
  \vspace{-10pt}
\end{figure*}

\begin{figure*}[t!]
  \centering
  \includegraphics[width=0.9\textwidth]{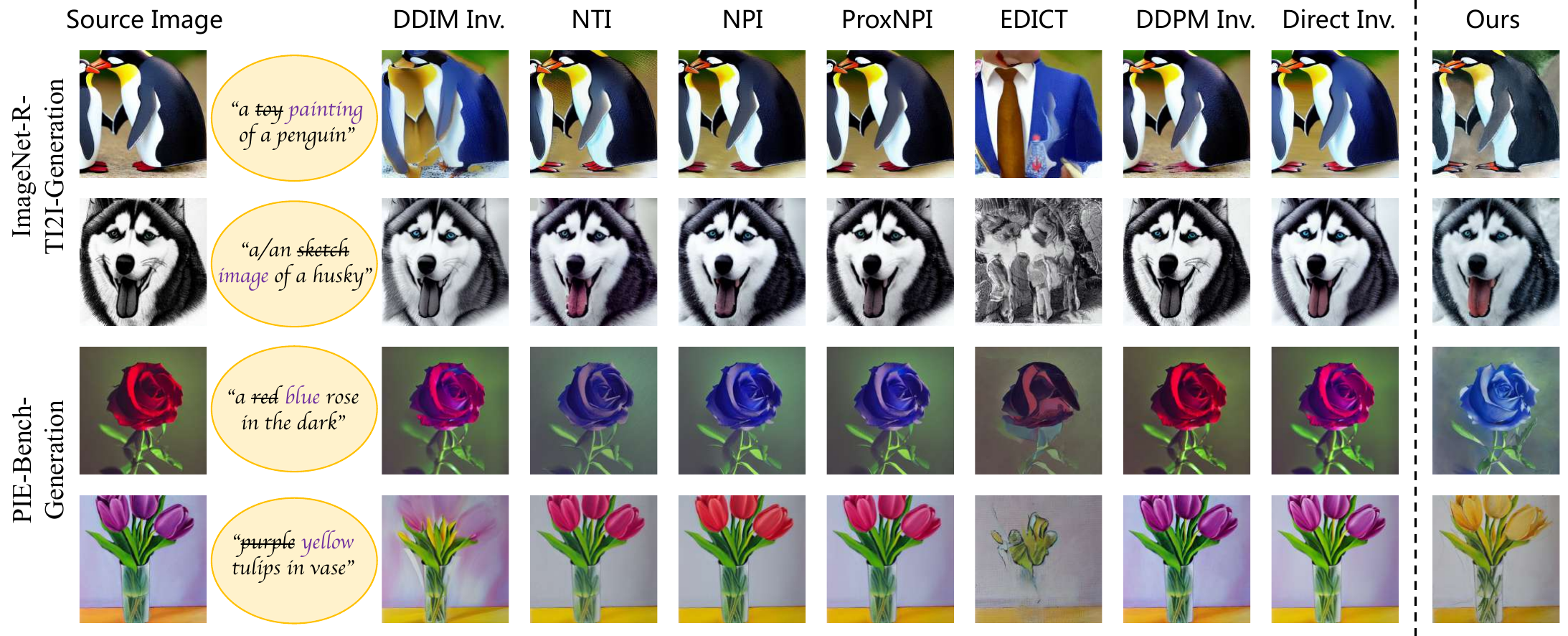}
  \caption{Qualitative comparison of different inversion methods combined with PnP on \textit{ImageNet-R-TI2I} and \textit{PIE-Bench}.}
  \label{fig:fig_5}
  \vspace{-10pt}
\end{figure*}

\begin{figure*}[t!]
  \centering
  \includegraphics[width=0.9\textwidth]{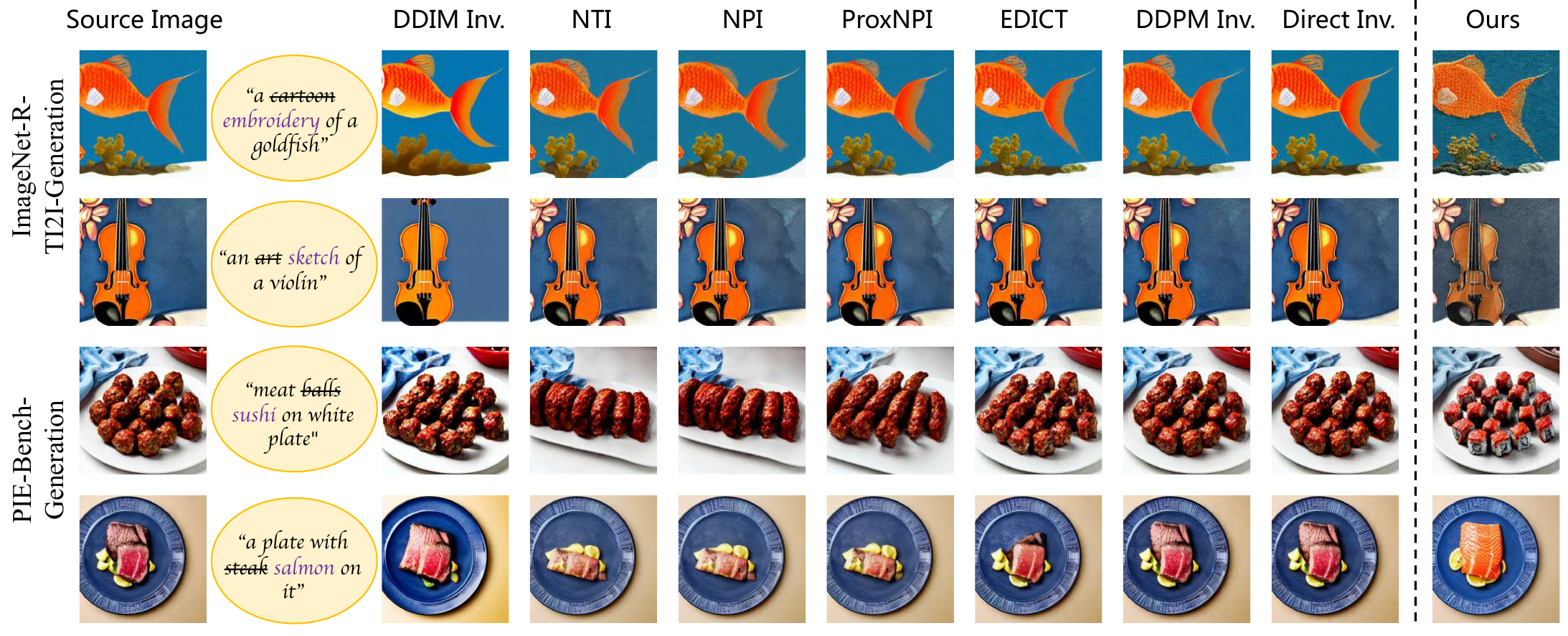}
  \caption{Qualitative comparison of different inversion methods combined with MasaCtrl on \textit{ImageNet-R-TI2I} and \textit{PIE-Bench}.}
  \label{fig:fig_6}
\end{figure*}

\subsection{Comparisons to Baselines}
\vspace{0.4cm}

We evaluate ResetEdit against a comprehensive set of inversion-based methods on two widely used datasets, \textit{PIE-Bench} and \textit{ImageNet-R-TI2I}. For each inversion method, we further combine three representative tuning-free editing pipelines—P2P, PnP, and MasaCtrl—yielding a two-dimensional comparison that jointly assesses inversion fidelity and editing controllability. The quantitative results are summarized in Table~\ref{tab:tab_1}.

Across both datasets, ResetEdit consistently outperforms all baselines in terms of text–image alignment and structural preservation. On \textit{PIE-Bench}, it improves CLIP Score by 1.2\%–2.5\% and achieves notable gains in CLIP Similarity, demonstrating that the reconstructed starting latent provides a more faithful anchor for aligning edits with target prompts. Similar improvements are observed on \textit{ImageNet-R-TI2I}, where ResetEdit achieves increases of 1.0\%–2.3\% in CLIP Score and 1.5\%–3.2\% in CLIP Similarity. These consistent gains across datasets and pipelines underscore the robustness of ResetEdit in enabling high-fidelity and semantically precise edits.

To provide a more fine-grained analysis, we further decompose the results on \textit{PIE-Bench} into specific editing categories under the three pipelines, as shown in Table~\ref{tab:pie_bench_all}. Across all methods, we observe that object replacement and addition are relatively easier tasks, yielding consistently higher CLIP Scores and directional similarities, whereas attribute modifications—particularly pose and material—remain the most challenging, with all inversion-based baselines exhibiting noticeable degradation. Style-level edits such as background and style transfer also reveal difficulties for most methods, where semantic misalignment and over-editing are common. ResetEdit consistently surpasses baselines across nearly all sub-categories. Notably, the gains are most pronounced in challenging attribute edits (e.g., pose and material), where ResetEdit improves both semantic alignment and structural consistency by a large margin. For object-level edits, our method not only enhances accuracy in replacements and additions but also mitigates undesired artifacts such as incomplete deletions. In style-level edits, ResetEdit achieves the highest robustness in background and global style changes, demonstrating its ability to balance semantic transfer with content preservation.  

Qualitative comparisons further validate these findings (Figures \ref{fig:fig_4}, \ref{fig:fig_5}, and \ref{fig:fig_6}). When combined with P2P, PnP, or MasaCtrl, ResetEdit reconstructs a reliable starting latent that establishes a more effective editing space. As a result, edits produced with ResetEdit are semantically faithful and structurally consistent, preserving object attributes and spatial layouts while avoiding artifacts. In contrast, their DDIM-inversion counterparts frequently yield misaligned edits and unwanted alterations. For instance, in Fig.\ref{fig:fig_5}, when transforming “a toy painting of a penguin”, existing inversion methods either fail to capture the intended painting style or introduce severe artifacts, whereas ResetEdit faithfully conveys the artistic style while preserving the penguin’s structure and pose. Likewise, in Fig.\ref{fig:fig_6}, when replacing “steak” with “salmon”, most baselines retain much of the original object appearance, only making superficial changes in texture or color. In contrast, ResetEdit achieves a complete and realistic replacement, while maintaining consistency in the plate and background. These examples highlight ResetEdit’s advantage in both style transfer and object-level replacement tasks, demonstrating its broad applicability for diverse editing scenarios.

\subsection{Ablation Study}

\begin{figure}[t!]
  \centering
  \includegraphics[width=\linewidth]{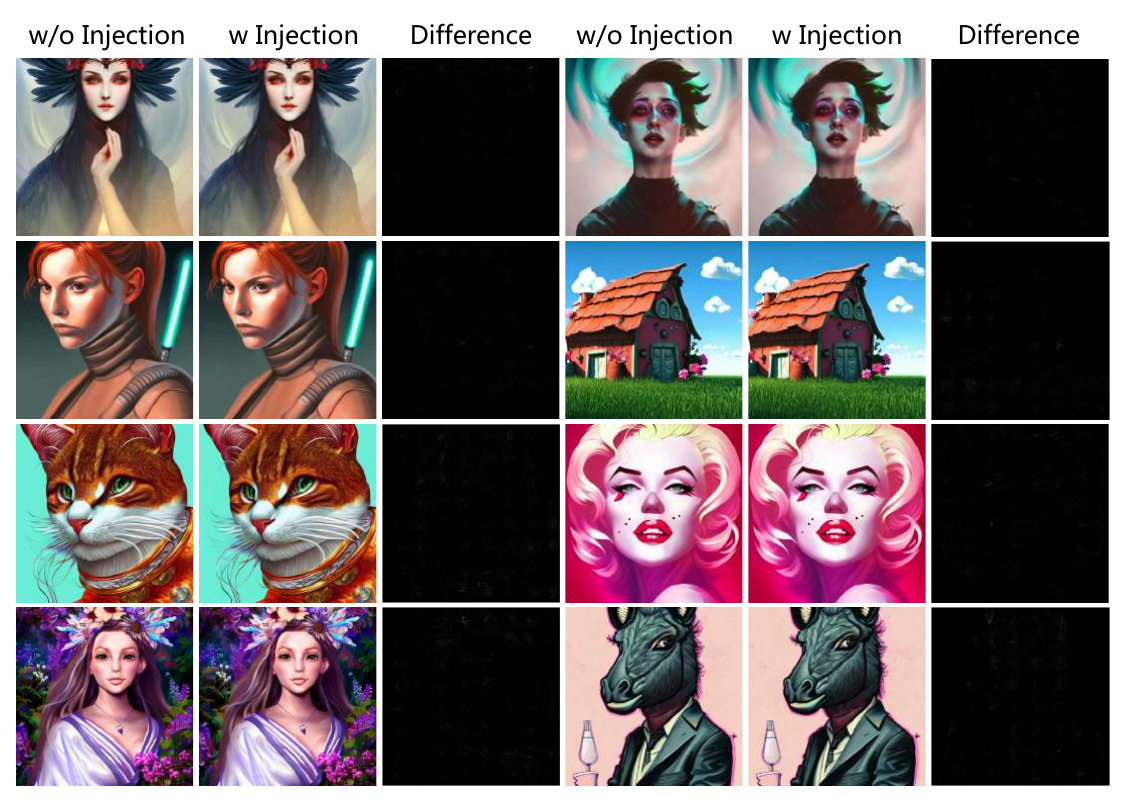}
  \caption{Effect of residual injection on image generation quality. From left to right: image generated without residual injection, image with injected residual, and their corresponding difference map. The average MSE is $8.47 \times 10^{-5}$, with a PSNR of 40.72\,dB, indicating high visual fidelity.}
  \label{fig:fig_7}
\end{figure}

\begin{figure}[t!]
  \centering
  \includegraphics[width=\linewidth]{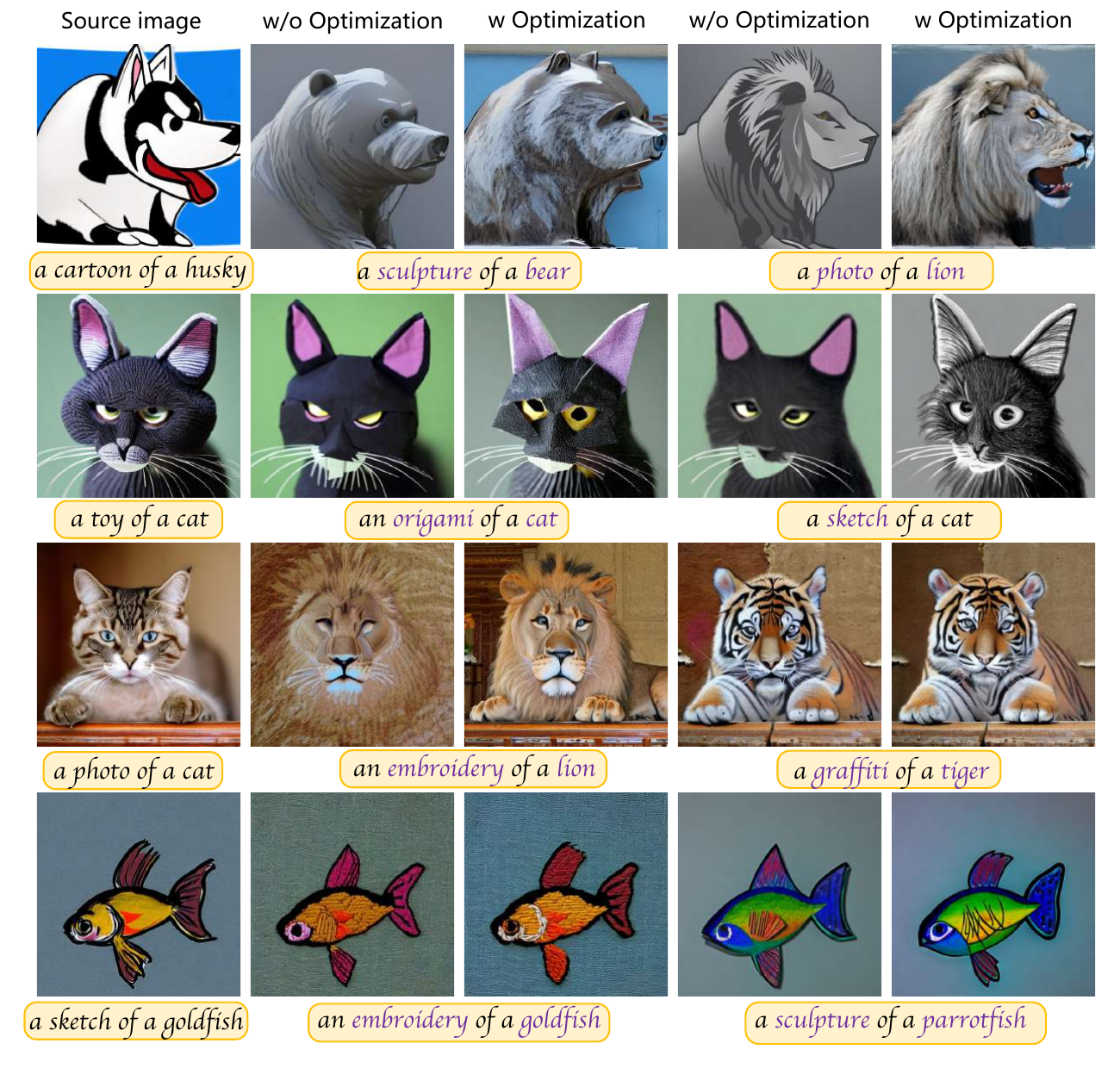}
  \caption{Effect of VAE optimization under the P2P \cite{hertz2022prompt} framework. From left to right: source image, edited result using the starting latent without VAE optimization, and edited result using the starting latent with VAE optimization.}
  \label{fig:fig_8}
\end{figure}

\noindent
\textbf{Effect of Injection on Image Generation Quality.}
To enable accurate recovery of the starting latent, we inject the residual into the diffused image latent prior to decoding. In this study, we examine whether this additional injection affects the visual quality of the generated images. As illustrated in Figure \ref{fig:fig_7}, we compare images generated with and without residual injection, along with their corresponding difference maps. The results show that the injected version maintains high visual quality, with minimal perceptual differences. This is largely due to the adaptive nature of the injection, which tends to insert information into visually complex regions such as textures, thereby preserving perceptual transparency while still enabling high-fidelity image reconstruction. Furthermore, the PSNR between the two versions reaches 40.72dB, quantitatively confirming that the embedding introduces negligible degradation.

\noindent
\textbf{Effect of VAE Optimization on Image Editing.}
To obtain a more accurate inversion of the starting latent, we further optimize the VAE encoding error before extracting the residual embedding. Specifically, we impose a constraint on the MSE between the input image and the reconstructed image, which allows us to refine the latent representation $z_0$,  providing a more precise initialization point for the inversion process. Due to the inherent robustness of the injection model we trained, the bit accuracy is consistently 1.0000 in all cases. However, when the VAE optimization is not applied, the starting latent is slightly misaligned, leading to a small increase in MSE. By applying just 20 iterations of VAE optimization, we achieve a more accurate starting latent, resulting in a significant reduction in MSE.  We visualize the impact of applying or omitting VAE optimization on image editing results in Figure \ref{fig:fig_8}. As illustrated, optimization leads to noticeably improved fidelity and consistency between the edited image and the intended target.

\section{Conclusion}
In this paper, targeting the task of post-generation image editing, where users refine generated images by modifying specific regions while preserving global structure, we propose ResetEdit, a novel editing framework featuring a resettable starting latent. By proactively injecting residual discrepancy signals during the forward process, our approach enables precise and controllable editing of generated images. To further reduce inversion errors, we optimize the image latent to ensure better alignment between the decoded and original images. Built on publicly available models, our method is compatible with existing editing techniques and achieves consistent improvements across datasets. Beyond editing, it also extends naturally to tasks such as inversion-based generative image watermarking, highlighting its broader applicability.

\bibliographystyle{ACM-Reference-Format}
\bibliography{reference}


\end{document}